\documentclass[runningheads]{llncs}
\usepackage{graphicx}
\usepackage{fancyhdr}
\usepackage{multirow}
\newcommand{\repeatthanks}{\textsuperscript{\thefootnote}}
\begin{document}
\title{A Survey on Complex Question Answering over Knowledge Base: Recent Advances and Challenges}
%
%
\author{Bin Fu\inst{1}\thanks{Both authors contributed equally to this work.} \and
Yunqi Qiu\inst{1,2}\repeatthanks \and
Chengguang Tang\inst{1} \and 
Yang Li\inst{1} \and  
Haiyang Yu\inst{1} \and 
Jian Sun\inst{1}}
\authorrunning{B. Fu et al.}
\pagestyle{fancy} 
\fancyhead{}

%
\institute{Alibaba Group, Beijing \and
Institute of Computing Technology, Chinese Academy of Sciences, Beijing
\email{\{bingo.fb,chengguang.tcg,ly200170,yifei.yhy,jian.sun\}@alibaba-inc.com}\\
\email{qiuyunqi19b@ict.ac.cn}}
\maketitle              

\begin{abstract}
Question Answering (QA) over Knowledge Base (KB) aims to automatically answer natural language questions via well-structured relation information between entities stored in knowledge bases.
In order to make KBQA more applicable in actual scenarios,
researchers have shifted their attention from simple questions to complex questions,
which require more KB triples and constraint inference.
In this paper, we introduce the recent advances in complex QA.
Besides traditional methods relying on templates and rules,
the research is categorized into a taxonomy that contains two main branches, 
namely Information Retrieval-based and Neural Semantic Parsing-based.
After describing the methods of these branches, 
we analyze directions for future research and introduce the models proposed by the Alime team.

\end{abstract}
\section{Background}
Question Answering (QA) over Knowledge Base (KB) uses rich semantic information to deeply understand natural language questions and provide answers from knowledge bases. 
A simple example is the question ``How tall is Yao Ming?'', which relies on the KB triple (\textit{Yao Ming}, \textit{height}, \textit{2.26m}). 
Since KBQA can benefit a variety of applications, 
it has attracted extensive attention from academic and industrial circles in recent years. 
Models developed by the Alime team have been widely used in telecommunications, insurance, taxation, and other fields. 

Although simple questions can be solved easily, KBQA still faces challenges when handling complex questions in real customer service scenarios where users tend to express specific information and require answers urgently. 
Generally, these questions include three types: 
(1) Questions under specific conditions, e.g., ``A taxpayer with a quarter sales volume of less than RMB 300,000 needs to issue a special value-added tax (VAT) invoice of RMB 50,000. Does the taxpayer need to pay additional taxes?'' 
(2) Questions with more intentions, e.g., ``Introduce mobile large-traffic and ultimate traffic packages.'' 
(3) Questions requiring constraint inference, e.g., ``Which is the cheapest 5G package that you have?''.

To cope with these complex questions, we have investigated the recent advances in KBQA. This paper first introduces mainstream KBQA datasets, and then describes several branches of methods applied to these datasets. 
Finally, we analyze the frontier trend in KBQA, and introduce the models developed by the Alime team.
  
\section{Dataset Introduction}
With the development of KBQA technologies, simple questions can be answered well, and the research focus has changed to complex QA. Figure~\ref{tab1} shows some attributes of complex QA datasets and introduces three relevant KB from the general domain.

\begin{table}
\centering
\caption{Several KBQA benchmark datasets involving complex questions.}
\label{tab1}
\setlength{\tabcolsep}{2mm}
\begin{tabular}{|l|l|r|c|}
\hline
Dataset &  Background KB & Size & Logical forms\\
\hline
WebQuestions~\cite{webq} & Freebase & 5,810 & No \\
ComplexQuestions~\cite{compq} & Freebase & 2,100 & No \\
WebQuestionsSP~\cite{webqsp} & Freebase & 4,737 & Yes \\
ComplexWebQuestions~\cite{compwebq} & Freebase & 34,689 & Yes\\
QALD & DBpedia & 50-500 & Yes\\
LC-QuAD~\cite{lcquad} & DBpedia & 5,000 & Yes\\
LC-QuAD 2.0~\cite{lcquad2} & DBpedia, Wikidata & 30,000 & Yes\\
\hline
\end{tabular}
\end{table}

\subsection{WebQuestions and Its Derivative Datasets} 
WebQuestions~\cite{webq} is constructed to solve real questions. 
Its questions are fetched from the Google Suggest API, 
and answers are annotated with the help of Amazon Mechanic Turk. 
This dataset is the most popular benchmark dataset. 
However, it still has two shortcomings:
\begin{itemize}
\item Only answers are labeled for the questions, and no logical forms are provided.
\item Simple questions account for 84\% of all questions, and only a few of them require multi-hop reasoning and constraint inference.
\end{itemize}

To address the first problem, WebQuestionsSP~\cite{webqsp} is proposed, in which the SPARQL query statements are annotated for questions in WebQuestions, 
and some questions are removed due to ambiguous expressions, unclear intentions, or no clear answers.
To address the second problem, ComplexQuestions~\cite{compq} introduces more questions that contain entity or type constraints, explicit or implicit time constraints, and aggregation constraints. 
ComplexWebQuestions~\cite{compwebq} modifies the SPARQL queries from WebQuestionsSP by including more constraints, and then generated natural language questions with the help of templates and Amazon Mechanic Turk.

\subsection{QALD Series}
Question Answering over Linked Data (QALD) is an evaluation subtask on the Conference and Labs of the Evaluation Forum (CLEF). 
It has been held annually since 2011 and provides several training sets and test sets each time. 
Among these questions,  complex ones account for about 38\%, 
including questions with multiple relationships and entities, 
such as ``Which buildings in art deco style did Shreve, Lamb and Harmon design?'', 
and questions containing time, comparative, superlative, and inference constraints, 
such as ``How old was Steve Jobs's sister when she first met him?''

\subsection{LC-QuAD}
In 2017, Trivedi et al.~\cite{lcquad} published a dataset based on DBpedia, namely Large-Scale Complex Question Answering
Dataset (LC-QuAD), 
and complex questions account for 82\% in the dataset, 
e.g., ``What are the mascots of the teams participating in the Turkish handball super league?'' 
To construct these complex questions,  
predefined SPARQL templates were firstly filled with seed entities and associated relations to generate specific SPARQL queries on DBpedia. 
Then, these queries were converted to natural language questions through  predefined question templates and crowdsourcing. 
Follow this framework, Dubey et al.~\cite{lcquad2} constructed a larger and more diverse KBQA dataset, namely LcQuAD 2.0, which contains more types of complex questions and is based on both Wikidata and DBpedia.

\section{Core KBQA Methods}
Traditional methods leverage manually defined templates and rules to parse complex questions. 
However, these methods demand researchers to be familiar with linguistic knowledge, and have limited scalability. 
Currently, with the progress of representation learning,
 there are two mainstream branches for KBQA methods: Information Retrieval-based (IR), and Neural Semantic Parsing-based (NSP). 

IR-based methods regard QA as binary classification or sorting over candidate answers. 
They firstly generate distributed representations of candidate answers and questions, 
and then calculate the matching score between the encoded answers and questions to select the final answer. 
Some of them leverage a framework of multi-hop reasoning to handle complex questions. 
These methods get rid of manually defined templates and rules, but lack of model interpretability. 
Moreover, they cannot handle complex questions requiring constraint inference.

Semantic Parsers aim to convert natural language questions to executable query languages. 
NSP-based methods construct Semantic Parsers based on neural networks to enhance the parsing capability and scalability. They usually map unstructured questions to structured logical forms, e.g. query graphs and high-level programming languages, which then are converted to executable queries through hand-crafted rules. 
Generally, the results of the NSP-based methods are slightly better than most of the IR-based methods.

\subsection{Traditional Methods}
Traditional methods mainly rely on predefined rules or templates to parse questions and obtain logical forms. 
Berant et al.~\cite{webq} implemented a standard bottom-up parser. 
First, they built a coarse mapping from question phrases to KB entities or relations using a KB and a large text corpus. 
Then, given a question, the proposed parser recursively constructs derivations based on a lexicon mapping question phrases to KB entities and relations, and four manually defined operations, including Join, Interaction, Aggregate, and Bridging. 
Meanwhile, the parser relies on a log-linear model over the hand-crafted features to guide itself away from the bad derivations and reduce the search space. 

\begin{figure}
\includegraphics[width=\textwidth]{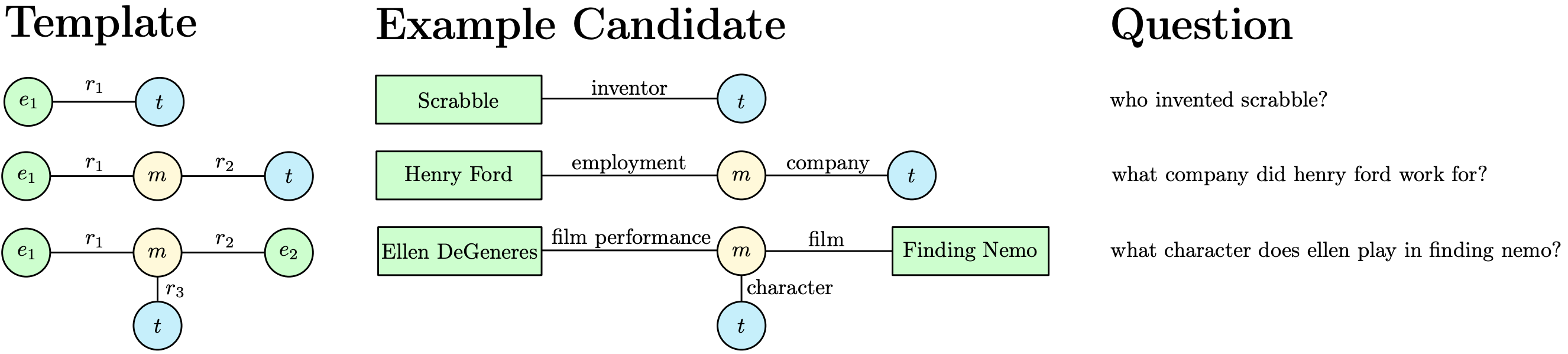}
\caption{Templates used to map questions in Aqqu and example candidates with corresponding questions.~\cite{bast2015}} 
\label{fig1}
\end{figure}

Bast et al.~\cite{bast2015} proposed a template-based model, namely Aqqu, which maps questions to three templates as shown in Figure~\ref{fig1}. 
At first, all entities that match a part of the question are identified from the KB. 
The match can be literal, or via an alias of the entity name. 
Then, Aqqu instantiates the three templates with the KB subgraph centered on the matched entity(s). 
According to the ranking model based on hand-crafted features, 
the best instantiation is output to query the KB and obtain the answers.

However, the three templates in Aqqu have limited coverage on complex questions. 
To handle more questions, some researchers try to automatically or semi-automatically learn templates from the KBQA datasets. 
Abujabal et al.~\cite{quint} proposed an automated template generation model, namely QUINT, 
and leveraged two kinds of templates, i.e., query template and question template. 
A query template is fetched from the KB through rules, 
and a question template is generated from the dependency parse result of the given question. 
During testing, a question is first mapped to some question templates, 
and then the corresponding query templates are instantiated to candidate queries. 
Ranked by a random forest classifier, the top-one query is output to obtain the final answer. 

\subsection{Information Retrieval-Based Methods}
The IR-based branch first determines the entities of interest (i.e., topic entities) mentioned in natural language questions and links these entities to the KB. 
Then, it extracts topic-entity-centric subgraphs and treats all the nodes in the subgraphs as candidate answers. 
Based on the features extracted from the questions and candidate answers, 
it uses score functions to model their semantic relevance and predicts the final answers. 
Based on the feature representation technology, 
IR-based methods can be classified into those based on feature engineering and those based on representation learning.

\subsubsection{Feature Engineering}
Yao et al.~\cite{yao2014} analyzed the question syntax information and extracted four types of features from each question’s dependency parse result, i.e., the question word (qword), question focus word (qfocus), topic word (qtopic), and central verb (qverb). All the four features are combined to form a question graph, and a KB subgraph is extracted according to the topic word. All the nodes in the KB subgraph are treated as candidate answers, 
then question features from the question graph and candidate answers’ features form the KB subgraph are combined with pairwise concatenation to capture the semantic association. 
Whether a node is a corresponding answer or not is determined by a classification model based on the above features. 
During training, high weights are assigned to features with high semantic associations.

\subsubsection{Representation Learning}
Methods based on feature engineering rely on manually defined and extracted features, 
which are time-consuming and cannot capture the whole semantic information of questions. 
To solve these problems, representation learning-based methods convert questions and candidate answers into vectors in the common vector space and treat KBQA as semantic matching calculation between distributed representations of questions and candidate answers. 
Some methods incorporate external knowledge to enrich the representations and handle the incompleteness of KB. 
Moreover, some methods leverage a framework of multi-hop reasoning to update these representations for complex questions.

\paragraph{One-hop reasoning}
Bordes et al.~\cite{bordes2014} first introduced the representation learning-based method.  
Embeddings of question words are summed as the question representation. 
Each candidate answer entity is represented by summing vectors from three aspects: the answer entity, the relationship path between the answer entity and the topic entity, and the subgraph related to the answer entity. 
The semantic relevance is calculated by dot product, and a margin ranking loss between positive and negative examples is used to train the model parameters.

With the progress of deep learning, neural networks are leveraged to generate better distributed representations for questions and candidate answers. 
Dong et al.~\cite{dong2015} proposed multi-column convolutional neural networks (MCCNNs) to encode questions from three different aspects. 
The proposed model learns three specific question representations corresponding to each aspect of answer features, i.e., answer type, answer path, and answer context (answer-centric subgraph), and different question representations capture different semantic information. 
The semantic score is the sum of the three dot-product results between the question representations and answer features. Compared with representing questions by bag-of-words, MCCNNs demonstrate the effectiveness of considering word order information and the relationship between questions and answers to improve the KBQA performance.

To capture more correlation information between questions and answers, 
Hao et al.~\cite{hao2017} proposed the cross-attention based neural network model to put more emphasis on question representation. 
Similar to previous methods, this model encodes candidate answers from four aspects: answer entity, answer path, answer type, and answer context. 
Then, the attention mechanism is leveraged to dynamically learn the correlation between different answer aspects and question words. 
In this way, the question learns the weights of different answer aspects, which effectively improves the QA performance. 

\paragraph{Incorporating External Knowledge}
Considering that the background knowledge bases are usually incomplete and previous methods are weak in constraint inference, 
some methods tend to incorporate external knowledge such as the web corpus and well-trained KB completion models. 
Xu et al.~\cite{xu2016} empowered a KBQA method with additional evidence from Wikipedia. 
At the first step, entity linking and relation extraction are performed to obtain the topic entity and potential KB relations mentioned in each question. 
After a joint inference over the entity linking and relation extraction results, 
candidate answer entities are obtained from the KB. 
Then, at the second step,  these candidate answers are refined by a binary classification model which takes the Wikipedia page of the topic entity into consideration to whether a candidate answer is positive or negative. 
This method is a practice of late fusion, and rely on the results of the first step, which may be incorrect. 

Instead of late fusion, 
Sun et al.~\cite{graftnet} proposed an early fusion strategy, where answers are extracted from a heterogeneous graph constructed from the KB and the corpus based on the given question. 
Specifically, the graph contains two types of nodes relevant to the question, i.e., entity nodes and sentence nodes. 
Besides KB relations between entity nodes, the graph contains one special type of edges which indicate entities are mentioned in the corresponding sentences. Then, the proposed graph convolution neural network, namely GRAFT-Net, will propagate feature information from one node to another. Finally, a node classification model determines whether an entity node in the graph is the answer or not. However, the question graph construction relies on heuristic rules and may result in error-cascading. Thus, Sun et al.~\cite{pullnet} proposed a learned iterative process for question graph construction. 
At each iteration step, seed entity nodes are selected by GRAFT-Net and a classification model. 
Then, with pre-defined operations, new entities and sentences are obtained from the KB and corpus, and added to the current graph. 
After several iterations, the answers are determined by the same way as~\cite{graftnet}. 

Previous methods all rely on the web corpus, 
but Saxena et al.~\cite{embedkgqa} assumed that such methods had limited coverage since it was not easy to get relevant corpus. 
They incorporated a well-trained KB completion model to handle KB incompleteness. 
Specifically, they first pre-trained a state-of-the-art KBC model, namely ComplEx~\cite{complex}, 
which leverages a score function to judge whether a pair of entities are connected by a specific relation. 
Then, the representations of the topic entity, the question, and the candidate answer are fed into the score function, 
and KBQA is also treated as a binary classification. 

\paragraph{Multi-hop reasoning}
Previous methods treat simple questions and complex questions in the same way. 
To acclimatize models to complex questions, more and more researchers adopt a framework of multi-hop reasoning. 
Among these methods, memory networks are widely used due to their excellent scalability and applicability to strong and weak supervision~\cite{wsmemnn}. 
Memory networks reason with inference components combined with a long-term memory component which can be read and written to, and stores KB triples~\cite{memnn,bordes2015}. 
Miller et al.~\cite{kvmemnn} proposed Key Value-Memory networks (KV-MemNN), 
which performs QA by first storing facts in a key-value structured memory before reasoning on them to predict an answer. 
They defined three operations, i.e., key hashing, key addressing, and value reading. 
During key hashing, KB triples relevant to the given questions are fetched, 
and the head entity and the relation are stored in the key slot, while the tail entity is stored in the value slot. 
During key addressing, each memory is assigned a normalized relevance weight by the dot product between the question representation and the key representation. 
During value reading, they take all the values’ weighted sum with the relevance weights, and the output vector indicates the intermediate reasoning state, which then is used to update the question representation. 
After repeating the key addressing and value reading several times, 
the final question representation is used to determine which candidate answer should be output.  
However, this method encodes questions and KB triples separately, and ignores the interaction between the two parts. 
Thus, Chen et al.~\cite{bamnet} proposed the bidirectional attentive memory network (BAMnet) model, 
which employs the attention mechanism to capture the correlation between questions and KB information, and uses the correlation to enhance question representations. 

Besides memory networks, path walk is another type of multi-hop reasoning method. 
Qiu et al.~\cite{srn} proposed a Stepwise Reasoning Network (SRN), which formulates QA as a sequential decision problem. 
The proposed model performs path walk over the KB to obtain the answer, and can be trained in an end-to-end manner with reinforcement learning. 
In their formulation, the agent is the learner and decision maker. 
At each of a sequence of discrete time steps, 
it performs path walk based on the state consisting of the given question, the topic entity, currently visited entity, and previous decisions. 
The set of candidate actions at each time step is comprised of all the connected relations and tail entities of the current visited entity. 
To capture the unique information of different parts of a question, SRN employs the attention mechanism to decide which part should be focused on at present.

In summary, IR-based methods get rid of a large number of manually defined templates, 
and can be trained in an end-to-end manner.
However, it cannot effectively handle complex questions, and most methods are weak in interpretability.

\subsection{Neural Semantic Parsing-based methods}
Methods based on Semantic Parsing usually convert natural languages into executable query languages. 
Compared with traditional SP-based methods, 
NSP-based methods construct Semantic Parsers based on neural networks to enhance the parsing capability and scalability, instead of relying on manually defined rules or templates. 
These methods usually map unstructured questions to intermediate logical forms (e.g., query graphs and trees), 
and further convert them into queries, such as SPARQL.

\subsubsection{Query Graph}
Recent works leverage graphs to represent questions, namely query graphs, which have strong representation ability and share topology commonalities with KB.  
Reddy et al.~\cite{graphparser} proposed GraphParser which takes advantage of the representational power of Combinatory Categorial Grammar (CCG) to parse questions. 
The proposed model creates ungrounded semantic graphs from CCG-derived semantic parses.  
Then these ungrounded semantic graphs are mapped to KB subgraphs through mapping edge labels to KB relations, type nodes to KB entity types, and entity nodes to KB entities. 
Note that math nodes remain unchanged, representing aggregation constraints. 
A beam search procedure is employed to find out the best semantic graph which is then converted to an executable query. GraphParser conceptualizes semantic parsing as a graph matching problem. 

Inspired by~\cite{graphparser}, Yih et al.~\cite{stagg} proposed a framework, namely Staged Query Graph Generation (STAGG). Different from GraphParser, the nodes and edges in the query graphs are closely resemble the exact entities and relations from the KB. 
In other words, a query graph is a restricted subset of lambda-calculus in graph representation. 
Thus, it can be straightforwardly translated into an executable query. 
A query graph consists of four types of nodes: grounded entities which are existing entities in the KB, existential variables which are ungrounded entities, lambda variable which represents the answer, and aggregation function which conducts numerical operations over other nodes. 
STAGG defines three stages to generate query graphs. 
First of all, an existing entity linking tool is employed to obtain candidate entities and their scores. 
Then, STAGG explores all the relationship paths between the topic entity and the answer node. 
To restrict the search space, it only explores paths of length 2 when the middle existential variable can be grounded to a CVT node\footnote{A compound value type (CVT) node is a kind of auxiliary nodes in Resource Description Framework (RDF) to maintain N-ary facts.} and paths of length 1 otherwise. 
All the paths will be scored by a deep convolutional neural network. 
Finally, constraint nodes are attached to the relationship path according to heuristic rules. 
At each of the three stages, a log-linear model is leveraged to score the current partial query graph, 
and the best final query graph is output to query the KB. 
STAGG effectively uses the KB information to crop the semantic parsing space, simplifying the difficulty of the task.

Bao et al.~\cite{compq} pointed out that STAGG cannot cover some complex constraints. 
Thus, they extended the constraint types and operators based on STAGG, including type constraints and explicit and implicit time constraints, and systematically proposed Multiple Constraint Query Graph (MultiCG) to solve these complex questions.
Generally, MultiCG still follows the framework of STAGG and only provides more rules to cover the complex questions mentioned in~\cite{compq}.

Yu et al.~\cite{yu2017} assumed that poor entity linking results pull down the QA performance. 
To improve the recognition accuracy, they integrated entity linking and relationship path identification to make the two components enhance each other.  
Specifically, they proposed the Hierarchical Residual BiLSTM (HR-BiLSTM) to encode questions and all relationship paths associated with the candidate topic entities in word-level and phrase-level, and calculated the similarity scores for all the questions. Only candidate topic entities connected to those highly-scored relations will be reserved. 
When handling constraints in questions, HR-BiLSTM also follows the practice in STAGG.

Previous methods focus on the entity linking or constraint attachment stage, while some researchers emphasize the score function. Luo et al.~\cite{luo2018} assumed that each semantic component in the query graph conveys only partial information of the question, and is not directly comparable with the whole question. 
Meanwhile, existing methods cannot capture the compositional semantics, resulting from encoding different components separately. 
Thus, they encoded the query graphs and questions from both local and global perspectives. 
With the help of dependency parsing, they split a question into different parts corresponding to different semantic components in the query graph. 
These parts are fed into a bidirectional GRU separately to obtain vectors, which are gathered to generate the local representation of the given question through a max-pooling operation. 
And another bidirectional GRU is employed to encode the original question to generate the global representation. 
The final question representation is the sum of the local and global representations. 
Similarly, different semantic components are encoded separately, and gathered to obtain the final query graph representation through a max-pooling operation. 
The cosine similarity between the representations of the question and the query graph replaces the score calculated in the second stage in STAGG to help rank the query graph. 
Maheshwari et al.~\cite{g2019} conducted an empirical investigation of neural query graph ranking approaches, 
and proposed a slot-matching model based on the self-attention mechanism, which exploits the structure of query graphs by comparing its parts with different representations of the question. 
Zhu et al.~\cite{zhu2020} proposed the Tree-to-Seq method to take the order of entities and relationships into consideration, and encoded query graphs with tree-based LSTM. 

It is obvious that the existing framework relies on the results of the entity linking stage, but there are some questions including no topic entity, e.g., ``Who died in the same place they were born in?'' 
Hu et al.~\cite{ganswer} proposed a State-Transition Framework (STF), which is a combination of GraphParser and STAGG. 
At the first step, STF labels all the entities, types, variables in the question. 
Then, according to the dependency parsing result of the question, STF connects these nodes through predefined four atomic operations (i.e., Connect, Merge, Expand, Fold). 
Meanwhile, these edges are mapped to KB relations through a CNN-based relation matching model. 
Upon these four operations, STF generates a semantic query graph, which is then mapped to the KB subgraph through rules to obtain the query graph. 
STF overcomes some shortcomings of STAGG, while it still lacks the ability to handle questions with complex aggregations.

To restrict the search space, STAGG only explores relationship paths with limited length, make it impossible to handle multi-hop questions. Some researchers propose an iterative query graph generation framework. Bhutani et al.~\cite{cikm2019} assumed that a complex query graph is constructed by several partial queries that are generated by STAGG. 
And the query composition is determined by the augmented pointer networks~\cite{compwebq}. 
Lan et al.~\cite{mstagg} defined three actions to construct a query graph, i.e., extend, connect, and aggregate. 
Specifically, an extend action extends the relationship path by one more KB relation connected to the current endpoint node, 
a connect action links a grounded entity to the current relationship path, 
and an aggregate action attaches the detected aggregation function to the current path. 
All three types of actions are determined by the policy network. 
Once an action is selected, all the possible intermediate results will be scored, 
and only a few will be maintained. 
Based on reinforcement learning, the proposed model can be trained in an end-to-end manner.

\subsubsection{Encoder-Decoder Method}
In addition to query graphs, many researchers leverage trees or high-level programming languages to represent natural language questions. 

Dong et al.~\cite{dong2016} proposed an enhanced encoder-decoder model based on the attention mechanism and reduced semantic parsing into a Seq-to-Seq problem. 
In addition to using a Seq-to-Seq model to convert questions to logical forms, 
they also proposed a Seq-to-Tree model that uses the hierarchical tree-structured decoder to capture the structure of logical forms. Xu et al.~\cite{xu2018} pointed out that the general sequence encoder ignores useful syntactic information. 
Thus, they leveraged a syntactic graph to represent word order, dependency, and constituency. 
Meanwhile, they used the Graph-to-Seq model, in which a graph encoder encodes the syntactic graph, and the recurrent neural network decodes the logical forms based on the state vector at each time step and the context information obtained through the attention mechanism. 
Although these methods achieve state-of-the-art performance without relying on manually predefined rules, they both require a large number of training annotations, which are rather expensive in most KBQA scenarios.

To get rid of strong supervision, Liang et al.~\cite{nsm} proposed the Neural Symbolic Machine (NSM), which is a Seq-to-Seq model trained with reinforcement learning. 
NSM converts the natural language questions into a logical program comprised of Lisp expressions. 
It is based on a Manager-Programmer-Computer framework, 
where the manager provides weak supervision through a reward indicating whether the program answers correctly, 
the programmer takes questions as input and decodes a bunch of Lisp expressions, 
and the computer executes these expressions to obtain the answers. 
Specifically, the computer is a Lisp interpreter, and provides code-assistance to the programmer to restrict the decoder vocabulary and ensures the validity of expressions generated by the programmer. 
Once the programmer decodes a valid Lisp expression, 
the computer will execute it and obtain an intermediate result from the KB. 
All the results are stored in a key-variable memory for the followup reasoning, 
and the final result is output as the predicted answer. 
The F1 score of the answer is the terminal reward for the programmer. 

In this section, we introduced the Neural Semantic Parsing-based methods, including methods that convert the Semantic Parsing into query graph generation, and methods leveraging an encoder-decoder framework. 
Although these methods can cover more complex questions, 
it is challenging to train a neural semantic parser, due to lack of gold logical forms.

\subsection{Other Methods}
In recent years, some new methods have been developed, which cannot be simply classified into the preceding types of methods. 
Talmor et al.~\cite{compwebq} split a complex question into several simple questions, 
each was submitted to a search engine, 
and then an answer was extracted from the search result with a Reading Comprehension model. 
The final answer could be computed with symbolic operations, such as union and intersection, over all the answers to the simple questions. 
Specifically, they used a Seq-to-Seq model to map complex questions to short programs that indicated how to decompose the question and compose the retrieved answers. 
To train the model, they performed a noisy alignment from machine-generated questions to natural language questions and automatically generated noisy supervision for training.

To equip traditional KV-MemNN with the ability to handle entity or type constraints,  
Xu et al.~\cite{enhancing} proposed the Enhancing KV-MemNN. 
During training, they employed the same learning method as the original KV-MemNN. 
While during testing, 
the proposed model selects the keys that have the highest relevance probabilities at every hop, 
and these selected keys are gathered to form a SPARQL query. 
To select keys correctly, Enhancing KV-MemNN introduces a new question representation updating strategy, 
taking both the addressed keys and addressed values into consideration.  
Meanwhile, it adds a special STOP key during memory readings, avoiding repeated or invalid memory readings.

\section{Frontier Trend Analysis}\label{fta}
Taking into account the current research work and the problems encountered in practical applications, 
we believe that future research will focus on the following three aspects.

\subsection{Represent Complex Questions}
Although there are many works in complex QA, some questions have more complex semantic structures that cannot be expressed by existing methods. 
Specifically, there are several types of questions not well studied in existing works: 
(1) Questions under specific conditions, e.g., ``A taxpayer with a quarter sales volume of less than RMB 300,000 needs to issue a special value-added tax (VAT) invoice of RMB 50,000. Does the taxpayer need to pay additional taxes?'' 
(2) Questions with more intentions, e.g., ``Introduce mobile large-traffic and ultimate traffic packages.'' 
(3) Questions requiring different logical connections, e.g., ``Which country participated in the World Cup finals most often?''. In this case, either the champions or the runner-ups should be taken account of.  
Therefore, it is of great significance to design a better logical form generation method to represent the semantic structures of complex questions. 
Specifically, the future works include improving the coverage of the generated logical forms and integrating KB information efficiently to restrict the search space.

\subsection{Enhance Model Robustness}
Jia et al.~\cite{jia2017} pointed out that in reading comprehension tasks, 
the model performance decreases dramatically when some irrelevant sentences are added to the paragraph. 
Similarly, natural language questions in KBQA are very colloquial, often with ambiguity or incorrect expressions. 
This situation brings great challenges to KBQA models since many papers show that entity linking and relation matching have always been important factors that affect the accuracy on complex questions. 
In actual applications, we found that many KBQA models are weak in the generation capability, and get trapped in the patterns in the training corpora, making it difficult to apply the KBQA models in industrial environments. 
Thus, enhancing the model robustness will be an important and promising trend for the industrial application of KBQA models.

\subsection{Multi-Turn Interaction}
In real scenarios, users usually ask multiple questions consecutively. 
Some questions can be answered by single-turn KBQA models, while some questions require models to leverage conversation context to resolve coreferences and ellipsis, and some ambiguous questions need clarifications from the users. However, KBQA and dialog systems have been studied independently. Recently, multi-turn QA has attracted more and more attention, but existing methods still have a lot of room for improvement. 
We believe that it is necessary to solve sequential questions in multi-turn QA for applications in real scenarios.

\section{KBQA of the Alime Conversational AI Team}
The Alime team has been continuously working on human-machine dialog methods and has built a KBQA engine applied in real business scenarios. The engine is faced with both simple questions and complex questions. Simple ones include ``Why can I not subscribe to the ultimate traffic package?'', ``Deferred tax payment'' and so on. The following table describes the different types of complex questions that the Alime KBQA needs to solve.

\begin{table}
\caption{Types of complex questions for the Alime KBQA.}
\label{tab2}
\centering
\begin{tabular}{|l|l|l|}
\hline
Question type &  Example \\
\hline
\multirow{5}{*}{Specific conditions}&Personal income tax for selling houses\\
    \cline{2-2}
    &How do I deal with the rebate and payment of personal income  \\ & tax if the taxpayer's income is in a foreign currency?\\
    \cline{2-2}
    & Mortgage was originally used as a personal income tax  \\ &deduction. Can it be changed to rent now?\\
\hline
\multirow{3}{*}{Temporal constraints}&How do I calculate the personal income tax for a year-end \\ & bonus issued after 2019?\\
    \cline{2-2}
    &Why can I not retrieve the personal income tax for 2019?\\
\hline
\multirow{2}{*}{Aggregation}&How much is the cheapest package?\\
\cline{2-2}
    &How much insurance can be received for accidents?\\
\hline
\multirow{4}{*}{Comparison}&What are the differences between common and special VAT \\ & invoices?\\
\cline{2-2}
    &What are the differences between corporate income tax, \\ & personal income tax, and individual business income tax?\\
\hline
\multirow{2}{*}{Boolean questions}&Do I have to pay personal income tax for an assessment bonus?\\
\cline{2-2}
    &I am 38 years old this year. Can I buy critical illness insurance?\\
\hline
\multirow{5}{*}{More intentions}&How much can children's accident insurance compensate at the  \\ & maximum? What is the maximum payment?\\
\cline{2-2}
    &I want to withdraw my insurance from Ping An Insurance. \\
    \cline{2-2}
    & What certificates should I take and where can I handle the  \\ & service?\\
\hline
\end{tabular}
\end{table}

In typical KBQA application fields (e.g., such as telecommunication, tax, and insurance), 
questions usually contain many entities and specific conditions. 
For example, in the telecommunication field, different methods are used to cancel packages (L) with different traffic (M) and prices (N). 
If using a FAQ method, we need to handle M x N x L knowledge points. 
As a result, the KB gets redundant, and users cannot obtain accurate answers. 
The KBQA solution we adopted can not only answer users' questions correctly, 
but also reduce manual cost to maintain a large FAQ KB.

\begin{figure}
\centering
\includegraphics[width=\textwidth]{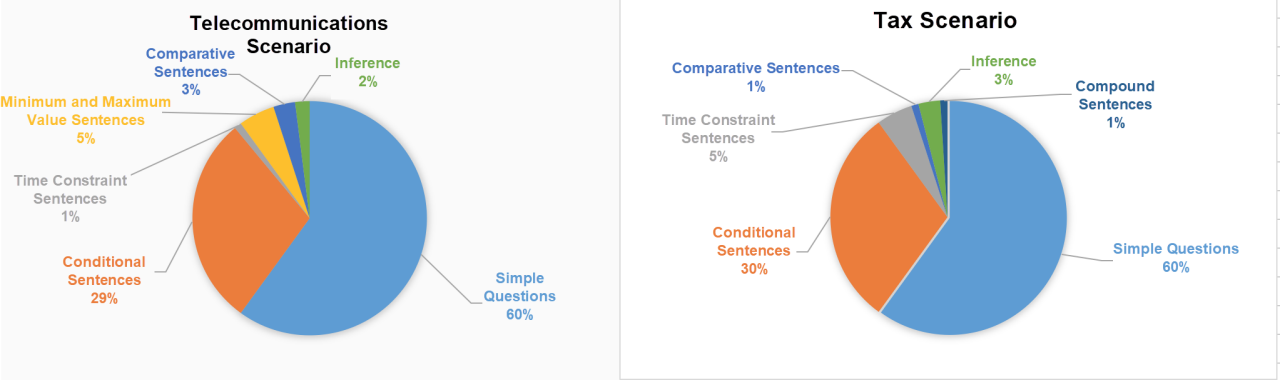}
\caption{Percentages of different types of KBQA questions in telecommunication and tax scenarios} 
\label{fig2}
\end{figure}

According to the data analysis, questions that can be answered with the help of KB account for 30\% to 40\% of all questions. Figure~\ref{fig2} shows the proportions of different types of these questions. 
Most questions are simple ones (60\%), which indicates that users tend to solve their real questions instead of testing the ability of the robot. 
However, complex questions requiring multi-hop reasoning and constraint inference also account for 40\%, which indicates the challenges for QA services in such scenarios. 
Thus, we focus on the method to accurately answer users' questions in these scenarios and optimize the users’ experience during human-machine interaction.

\begin{figure}
\centering
\includegraphics[width=\textwidth]{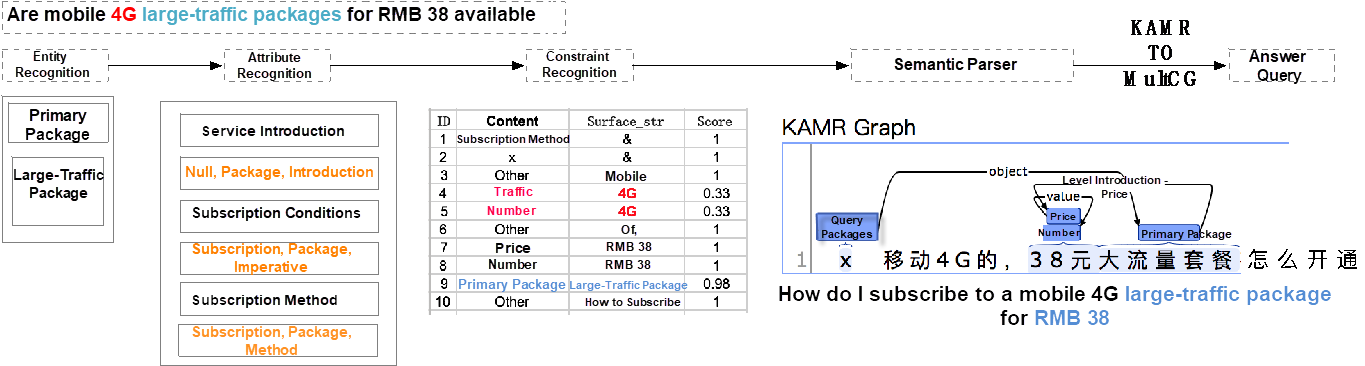}
\caption{MultiCG with dependency parser} 
\label{fig3}
\end{figure}

By building KB for vertical fields, we help enterprises mine the inherent structured information behind business knowledge, 
and turn scattered non-structured knowledge (graphite) into organic structured knowledge (diamonds) to improve the value of knowledge in vertical fields. 
Based on MultiCG, we have developed a Neural Semantic Parsing algorithm to support KBQA in vertical fields, as shown in Figure~\ref{fig3}. This algorithm uses the dependency parse results of the given questions to disambiguate constraints. In practical applications, the effective rate of this algorithm is greater than 80\%, and it also helps enterprises scale down their FAQ KB by nearly 10 times.

\begin{figure}
\centering
\includegraphics[width=\textwidth]{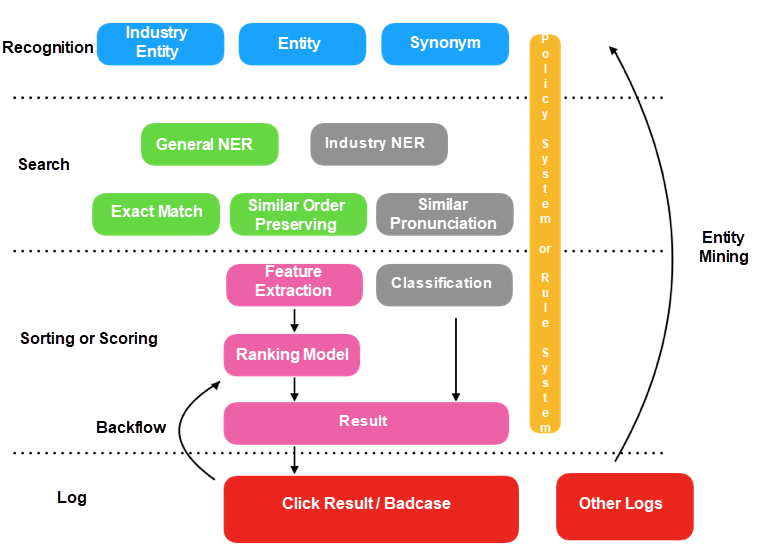}
\caption{Search and rank framework} 
\label{fig4}
\end{figure}

In Section~\ref{fta}, we have pointed out that it is urgent to solve entity linking and relation matching in real scenarios. 
In the entity linking step, entity disambiguation is not the focus because there are few entity types in the KB of vertical fields. However, discontinuous or nested entities may occur in user expressions. 
For example, the entity corresponding to ``Feixiang 4G package of RMB 18'' is ``Feixiang package''. 
To better identify the mentions in users' questions and link them to KB entities, 
we proposed a search and rank framework, as shown in Figure~\ref{fig4}. 
This framework uses matching and order-preserving similarity to match mentions and entities in a question, 
and then employs the ranking model to select the optimal entity or recommend entities.

\begin{figure}
\centering
\includegraphics[width=\textwidth]{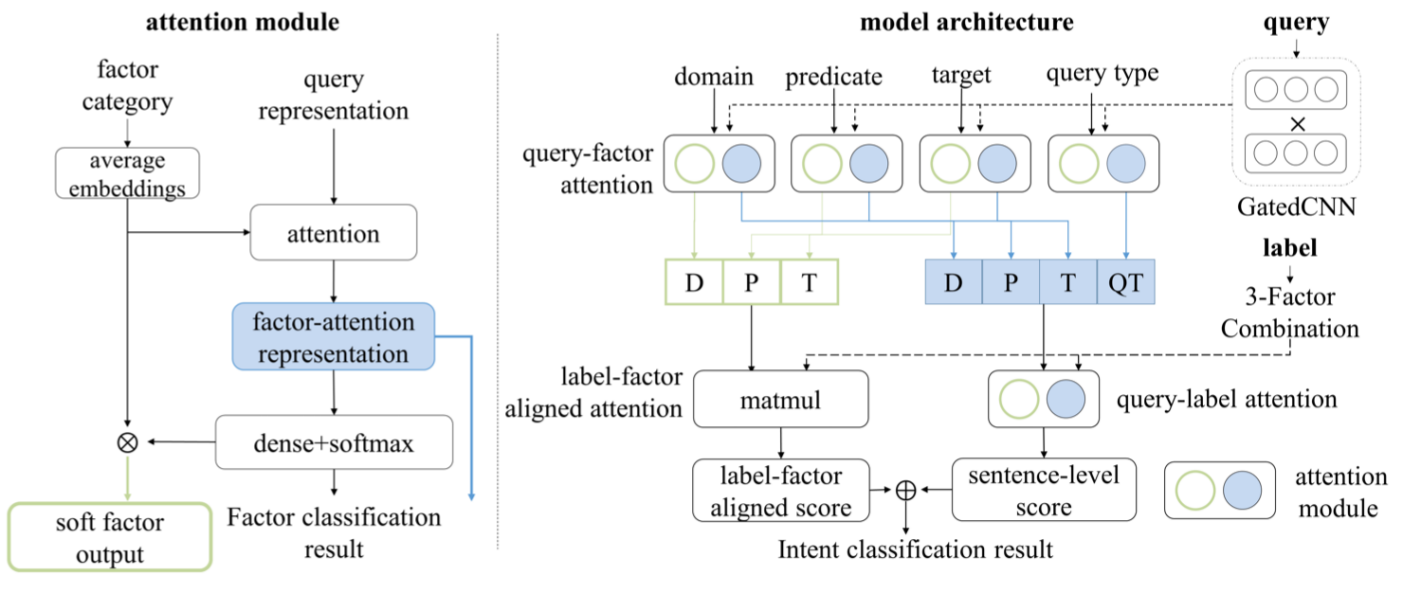}
\caption{CIBA model framework} 
\label{fig5}
\end{figure}

For relation matching, we proposed a multi-point semantic representation framework~\cite{zhang2020}, 
which breaks down each relation into four fine-grained factor information, 
i.e., the topic, predicate, object or condition, and query type to distinguish confusing relations. 
Then, it uses compositional intent bi-attention (CIBA) shown in Figure~\ref{fig5} to combine coarse-grained relation information and fine-grained factor information with question representation to enhance the semantic representation of questions.
The experimental results show that our method can reduce the number of incorrect confusing relations through relation factorization and improve the performance of overall classification.

To enhance the model robustness and get rid of manually defined rules during the constraint recognition, 
we proposed the hierarchical KB-attention model based on the KV-MemNN, 
and captured the relationship between core components of questions and enhance hierarchical semantic understanding of questions through fine-grained modeling of the interaction between questions and the KB. 
The hierarchical KB-attention model outperforms MultiCG and has better interpretability and stronger capability in parsing complex questions.

\section{Summary}
Question answering (QA) over knowledge bases (KB) has attracted wide attention from researchers and companies. This paper introduced the core challenge in KBQA, i.e., complex questions. Based on several widely-used benchmark datasets, we summarized the research progress of KBQA. In addition to traditional methods relying on templates or rules, there are two mainstream branches: (1) Information Retrieval-based (IR), and (2) Neural Semantic Parser-based (NSP).  In IR-based methods, we introduced some classic models which leverage feature engineering or representation learning. These methods reduce the dependency on manually defined templates or rules but are weak in the model interpretability. Meanwhile, they cannot handle complex questions requiring constraint inference. In NSP-based methods, we introduced two kinds of logical form generation models, i.e., query graph-based, and encoder-decoder-based. Taking into account the current research progress and the problems encountered in practical applications, we provided an analysis of future research trends. Finally, we briefly introduced the KBQA development progress of the Alibaba DAMO Academy Alime Conversational AI team and analyzed the types of complex questions in different scenarios. 

%
%
%
%

\end{document}